\documentclass[conference]{IEEEtran}
\IEEEoverridecommandlockouts
\usepackage{cite}
\usepackage{url}
\usepackage{amsmath,amssymb,amsfonts}
\usepackage{algorithmic}
\usepackage{graphicx}
\usepackage{textcomp}
\usepackage{xcolor}
\def\BibTeX{{\rm B\kern-.05em{\sc i\kern-.025em b}\kern-.08em
    T\kern-.1667em\lower.7ex\hbox{E}\kern-.125emX}}

\DeclareRobustCommand*{\IEEEauthorrefmark}[1]{%
	\raisebox{0pt}[0pt][0pt]{\textsuperscript{\footnotesize\ensuremath{#1}}}}

\title{Infrared and Visible Image Fusion with Hierarchical Human Perception}

\author{
	\IEEEauthorblockN{
		\\
		Guang Yang\IEEEauthorrefmark{1},
		Jie Li\IEEEauthorrefmark{1*},
		Xin Liu\IEEEauthorrefmark{1}, 
		Zhusi Zhong\IEEEauthorrefmark{1}, and
		Xinbo Gao\IEEEauthorrefmark{1,2}}
		\\
	\IEEEauthorblockA{\IEEEauthorrefmark{1}School of Electronic Engineering, Xidian University, Xi’an, China}
	\IEEEauthorblockA{\IEEEauthorrefmark{2}Key Laboratory of Image Cognition, Chongqing University of Posts and Telecommunications, Chongqing, China}
 }
\begin{document}
\maketitle

\begin{abstract}
Image fusion combines images from multiple domains into one image, containing complementary information from source domains. Existing methods take pixel intensity, texture and high-level vision task information as the standards to determine preservation of information, lacking enhancement for human perception. We introduce an image fusion method, Hierarchical Perception Fusion (HPFusion), which leverages Large Vision-Language Model to incorporate hierarchical human semantic priors, preserving complementary information that satisfies human visual system. We propose multiple questions that humans focus on when viewing an image pair, and answers are generated via the Large Vision-Language Model according to images. The texts of answers are encoded into the fusion network, and the optimization also aims to guide the human semantic distribution of the fused image more similarly to source images, exploring complementary information within the human perception domain. Extensive experiments demonstrate our HPFusoin can achieve high-quality fusion results both for information preservation and human visual enhancement.
\end{abstract}


\section{Introduction}
Image fusion is a kind of pixel-level multi-domain data fusion task that incorporates images from different sources into a single image. The generated high-quality fusion results should effectively preserve essential information from the input domains, while compressing redundant information, expected to be more informative for human perception\cite{li2017pixel}.

Infrared images can easily distinguish thermal targets from the background in bad weather conditions, but suffer from low resolution, while visible images usually contain rich texture but are sensitive to dark or bad conditions\cite{ma2019infrared}. Early infrared and visible image fusion (IVF) methods often treated infrared thermal information as pixel intensity, and texture information as gradients to constrain the fusion network to integrate complementary information\cite{li2018densefuse}. Some recent methods\cite{tang2022image, liu2023multi} cascade with high-level vision tasks to guide the fused images retaining features, which can improve results like detection and segmentation. Above IVF methods pursue higher statistical evaluation metrics and metrics of high-level vision tasks to satisfy human visual perception, but they barely consider language prior, which may result in fusion results not aligning with human subjective perceptions.

Recently, with the development of Large Vision-Language Model (LVM), image generation is capable of generating images that conform to human visual perception by aligning and integrating visual information with text information\cite{liang2023iterative, yang2023implicit, sun2024coser}. SUPIR\cite{yu2024scaling} adopts LLaVA\cite{liu2024visual} to generate textual descriptions as the multi-modality language guidance to restore images. 

In this paper, we propose a method named Hierarchical Perception Fusion (HPFusion) to introduce hierarchical human priors into the fusion network by utilizing guidance of LVM. We first propose multiple questions people tend to ask when viewing an infrared and visible image pair, such as 'What is the content of the image?' and 'What targets are significant in this image?'. These questions demonstrate a hierarchical semantic structure from the overall context to specific region, simulating human semantic perception when trying to comprehend the image. In our work, we set four question sets to ask, which is shown in Fig.~\ref{fig:qa}. Then, we use LLaVA to answer these questions according to the input image pairs, and the emphasis of answers differs between infrared images and visible images due to the modality characteristic. These textual prompts are encoded by the CLIP (Contrastive Language-Image Pre-training)\cite{radford2021learning} and merge with the fusion network to guide the fusion process to be more semantically and contextually. Lastly, the optimization of network combines the traditional fusion loss and hierarchical semantic loss in the CLIP space, with the latter being constrained by the CLIP text embeddings between the fused image and input image pair. 

\section{Related Work}
\subsection{Infrared and Visible Image Fusion}
Traditional methods adopt sparse representation \cite{veshki2022coupled} and multi-scale transform \cite{zhou2016perceptual} to handle information extraction and fusion. DenseFuse\cite{li2018densefuse} is the first deep learning (DL)--based infrared and visible image fusion method. FusionGAN\cite{ma2019fusiongan} utilizes generative adversarial network (GAN) to integrate thermal radiation and texture in an implicit way. TarDAL\cite{liu2022target} cascade image fusion and object detection to mine information beneficial for both human inspection and high-level tasks. DDFM\cite{zhao2023ddfm} formulates fusion task as a conditional generative problem under the denoising diffusion model. FILM\cite{zhao2024image} generates textual descriptions via the ChatGPT to convey a deep semantic understanding of the fusion network, guiding the extraction of crucial visual features.

\subsection{Large Visual-Language Model}
CLIP\cite{radford2021learning} learns from natural language supervision at large-scale tasks during pre-training, exhibiting robust transfer performance without fine-tuning on various tasks, which have been widely used in low-level vision tasks. BLIP\cite{li2022blip} bootstrap image captions from noisy web data, which performs substantial improvement on different vision-language tasks.  LLaVA\cite{liu2024visual} is an Large Multimodal Model (LLM) that connects visual encoder of CLIP with the large language models, and fine-tunes the model on their generated instructional vision-language data. It can be capable of multimodal chat abilities for visual and language understanding.

\section{Method}
In this section, we introduce the paradigm of our hierarchical human perception image fusion method.

\begin{figure}
\centerline{\includegraphics[width=3.6in]{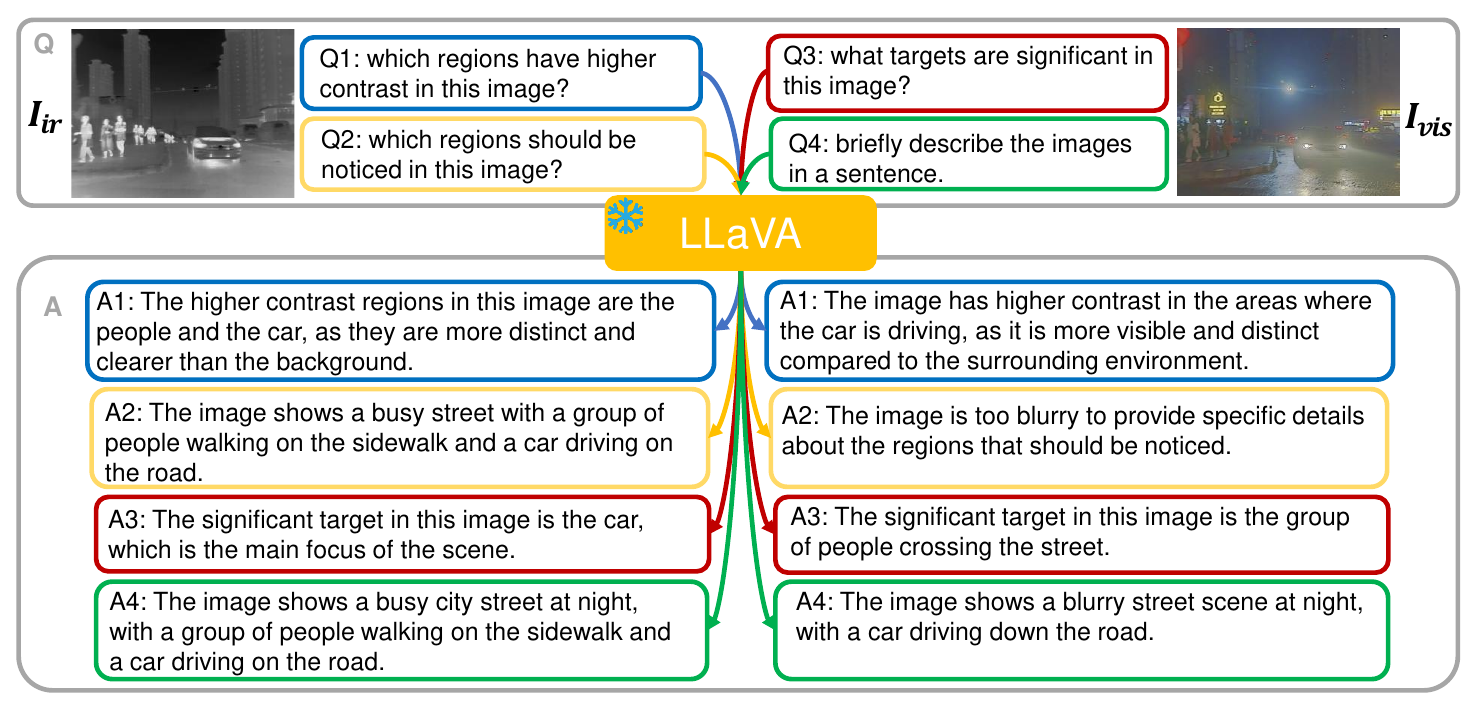}}
\caption{Questions that humans tend to ask when viewing the infrared and visible image pair and corresponding answers generated by LLaVA.}
\label{fig:qa}
\end{figure}

\subsection{Overview}
Infrared images can highlight the thermal targets and provide high contrast with the background where people can distinguish essential information. Visible images usually contain finer details, such as texts and signs, where people also tend to pay attention to. Meanwhile, the overall information describing an image is also mindful.

We have devised four questions from a global context to specific local regions, in order to simulate human curiosity and concerns when observing image pairs that are abundant in both thermal and detailed information. For example, when we view the infrared and visible image pair in Fig.~\ref{fig:qa}, we first attempt to comprehend the overall content so we ask LLaVA to describe the image (Q4). Subsequently, given that infrared images can usually highlight thermal targets, our objective is to identify the presence of various objects within the image (Q3). Lastly, infrared images have higher contrast than visible images, but visible images can also contain more details within a specific region. Therefore, we will focus on specific regions with high contrast and rich information content (Q1 and Q2). The generated textual prompts which incorporate hierarchical human priors are fused with the fusion network to guide the retention of useful visual features.

\subsection{Architecture}
\begin{figure*}[!t]
\centerline{\includegraphics[width=5.2in]{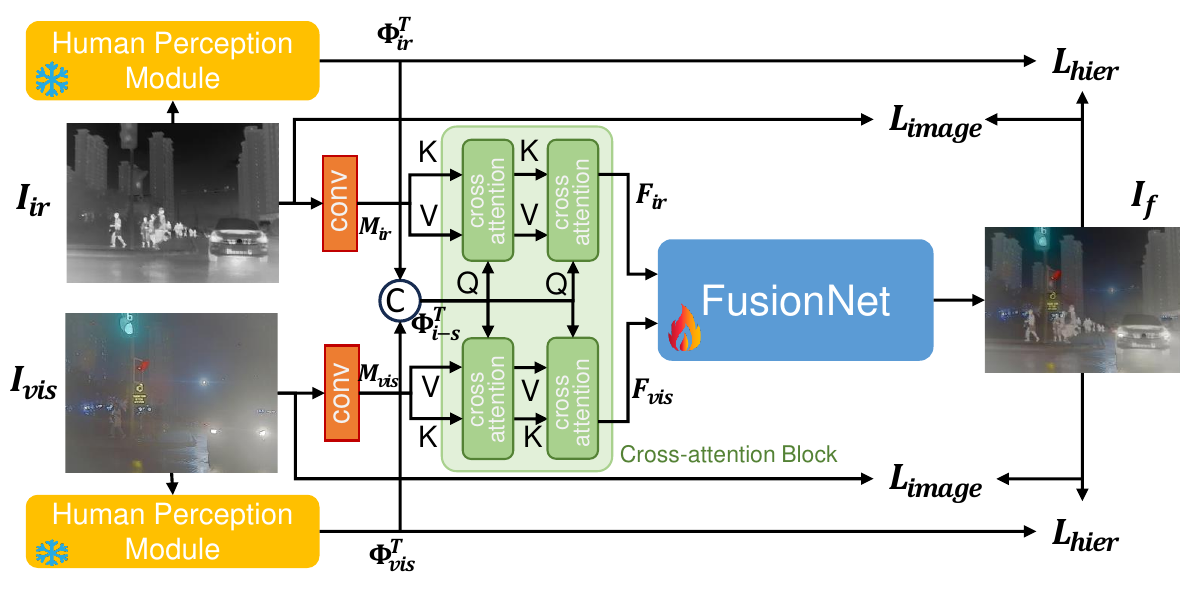}}
\caption{The overall architecture of our fusion network, consisting of Human Perception Module, Cross-attention Block and Fusion Network.}
\label{fig:architecture}
\end{figure*}

The overall architecture is shown in Fig.~\ref{fig:architecture}. For the given infrared and visible image $I_{ir}$ and $I_{vis}$, we design four questions that people tend to notice according to the thermal information of infrared and texture information of visible images to simulate hierarchical human perception.

Human Perception Module (HPM) consists of LLaVA and CLIP encoder to generate text features $\Phi_{ir}^T$ and $\Phi_{vis}^T$ as human priors. To guide the fusion of visual features, we use $1 \times 1$ convolutional layer to reduce the text features of the 4 answers to 1 dimension, then concatenate them to form the fused text feature $\Phi_{i-s}^T$.

The architecture of the fusion network follows the previous work MDA\cite{yang2024multi} as baseline, and the fused text feature $\Phi_{i-s}^T$ is combined with visual feature $M_{ir}$ and $M_{vis}$ extracted by the convolutional block of MDA, guiding the subsequent fusion of visual features, which is as follows:

\begin{equation}
\label{eq:cross_att}
  F_{ir}, F_{vis} = CA(M_{ir}, M_{vis}, \Phi_{i-s}^T)
\end{equation}
where $CA(\cdot)$ denotes the cross-attention mechanism, and the Query (Q) is calculated by the $\Phi_{i-s}^T$, while Key (K) and Value (V) are calculated by the $M_{ir}$ and $M_{vis}$. We cascade two Cross-attention Blocks to integrate the text and visual features comprehensively.

Finally, $F_{ir}$ and $F_{vis}$ are fused with the multi-scale encoder and decoder of MDA and then reconstruct the fused image $I_f$.

\begin{figure}
\centerline{\includegraphics[width=3.3in]{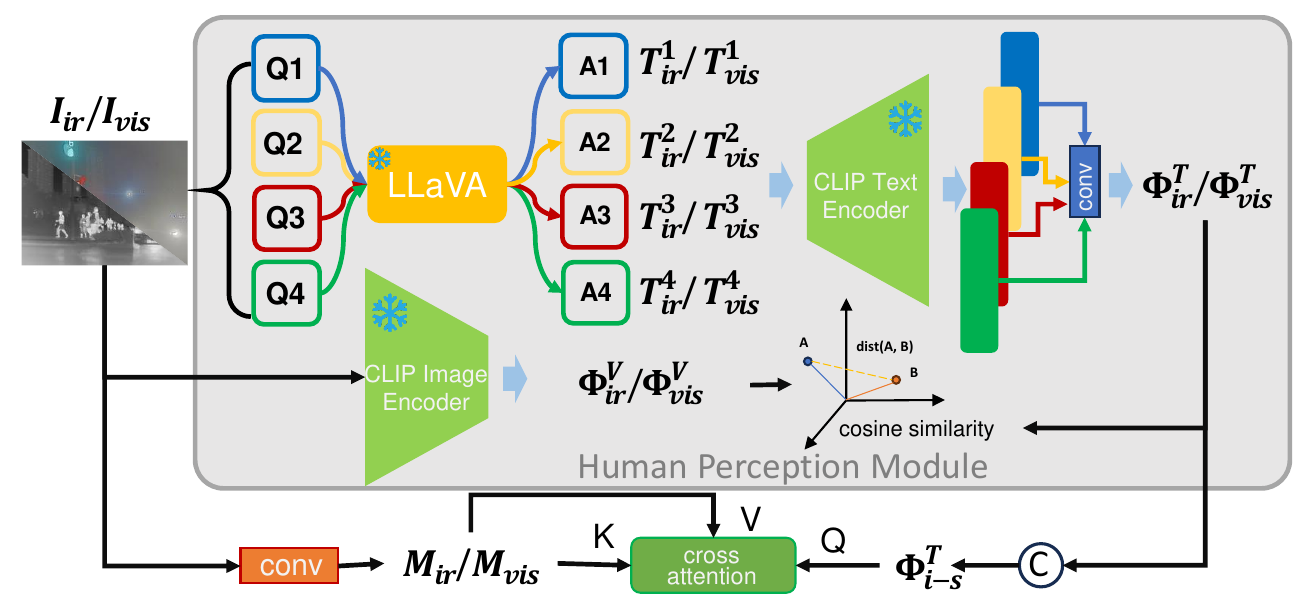}}
\caption{Architecture of the Human Perception Module.}
\label{fig:hier}
\end{figure}

\subsection{Human Perception Module}
The architecture of HPM is shown in Fig.~\ref{fig:hier}. For the input image pair, we ask four questions and answer via LLaVA according to infrared and visible images. After obtaining the text of answers $T_{ir} \in \mathbb R^{77 \times 4}$ and $T_{vis} \in \mathbb R^{77 \times 4}$ from LLaVA, we encode texts of these answers by parameter-frozen CLIP text encoder $\mathcal E_{text}$ to get the text features $\Phi_{ir}^T$ and $\Phi_{vis}^T$, guiding both fusion of visual features and optimization of training process. 

To guide the fusion of visual features, $\Phi_{ir}^T$ and $\Phi_{vis}^T$ are dimensionally reduced through the $1 \times 1$ convolutional layer and concatenation, and then $\Phi_{i-s}^T$ is fused with visual features  $M_{ir}$ and $M_{vis}$ by cross-attention mechanism. To guide the optimization of training process, the input image pair $I_{ir}$ and $I_{vis}$ are encoded by the parameter-frozen CLIP image encoder $\mathcal E_{img}$ to generate visual features $\Phi_{ir}^V$ and $\Phi_{vis}^V$, and text-image similarity inner the batch is calculated by the cosine similarity between $\Phi_{ir/vis}^V$ and $\Phi_{ir/vis}^T$ to represent semantic distribution.

Similarly, text-image similarity inner the batch of the fused image $I_f$ is calculated via the above method as well. Specifically, after obtaining the fused image $I_f$, we also ask the same four questions for $I_f$ and use LLaVA to generate the text of answers $T_f$, and encode answers into text features $\Phi_f^T$ by CLIP text encoder $\mathcal E_{text}$. Then the cosine similarity can be calculated by the $\Phi_f^T$ and $\Phi_f^V$.

\subsection{Loss Function}
The degree of information retention is determined by the optimization objective, thus we construct image loss $L_{image}$ to preserve both intensity and detail information from source images, while hierarchical semantic loss $L_{hier}$ aims at driving texts of answers guiding the preservation of information that prioritizes human focus and comprehension.

The image loss includes pixel intensity and detail items to maintain informational fidelity, which are maximum intensity loss, maximum gradient loss and structural similarity index metric (SSIM) loss, as shown in \eqref{eq:int_loss} and \eqref{eq:detail_loss}.
\begin{equation}
\label{eq:int_loss}
  L_{int} = {\left \| I_f - max(I_{ir}, I_{vis}) \right \|}_1
\end{equation}
\begin{equation}
\label{eq:detail_loss}
\begin{split} 
  L_{detail} &=   (1 - SSIM(I_f, I_{ir})) + (1 - SSIM(I_f, I_{vis})) \\
    &+ {\left \| \nabla I_f - max(\nabla I_{ir}, \nabla I_{vis}) \right \|}_1
\end{split} 
\end{equation}

The hierarchical semantic loss constrains the salient regions that humans focus on between fused image and source image pair. We expect the text-image similarity between the fused image and source image pair to be close, indicating that the distribution of each image with its textual answers is similar between the fused image and source image pair. Firstly, we measure the similarity score between the text vector of an answer and the corresponding image vector by calculating the cosine similarity in the CLIP space, which is as follows:
\begin{equation}
\label{eq:clip_loss_item}
   S(I_m) = \frac{e^{cos(\mathcal E_{img}(I_m), \mathcal E_{text}(T_m^i))}}{\sum_{i \in \{ batch \}} e^{cos(\mathcal E_{img}(I_m), \mathcal E_{text}(T_m^i))}}
\end{equation}
where $m \in \{ ir, vis, fusion \}$, and $i$ denotes other samples within a batch. $\mathcal E_{img}$ and $\mathcal E_{text}$ are parameter-frozen CLIP image and text encoder, respectively. Then, we adopt the similarity score between the fused image and the source image pair to be closer for the four answers, which expressed as hierarchical human concerns:
\begin{equation}
\label{eq:clip_loss}
    L_{hier} = \sum_{j \in 4} \left( \left \| S(I_f^j) - S(I_{ir}^j) \right \|_1 + \left \| S(I_f^j) - S(I_{vis}^j) \right \|_1 \right)
\end{equation}
where $j$ corresponds to four different question-answer sets.
\begin{figure*}[htbp]
\centerline{\includegraphics[width=6.0in]{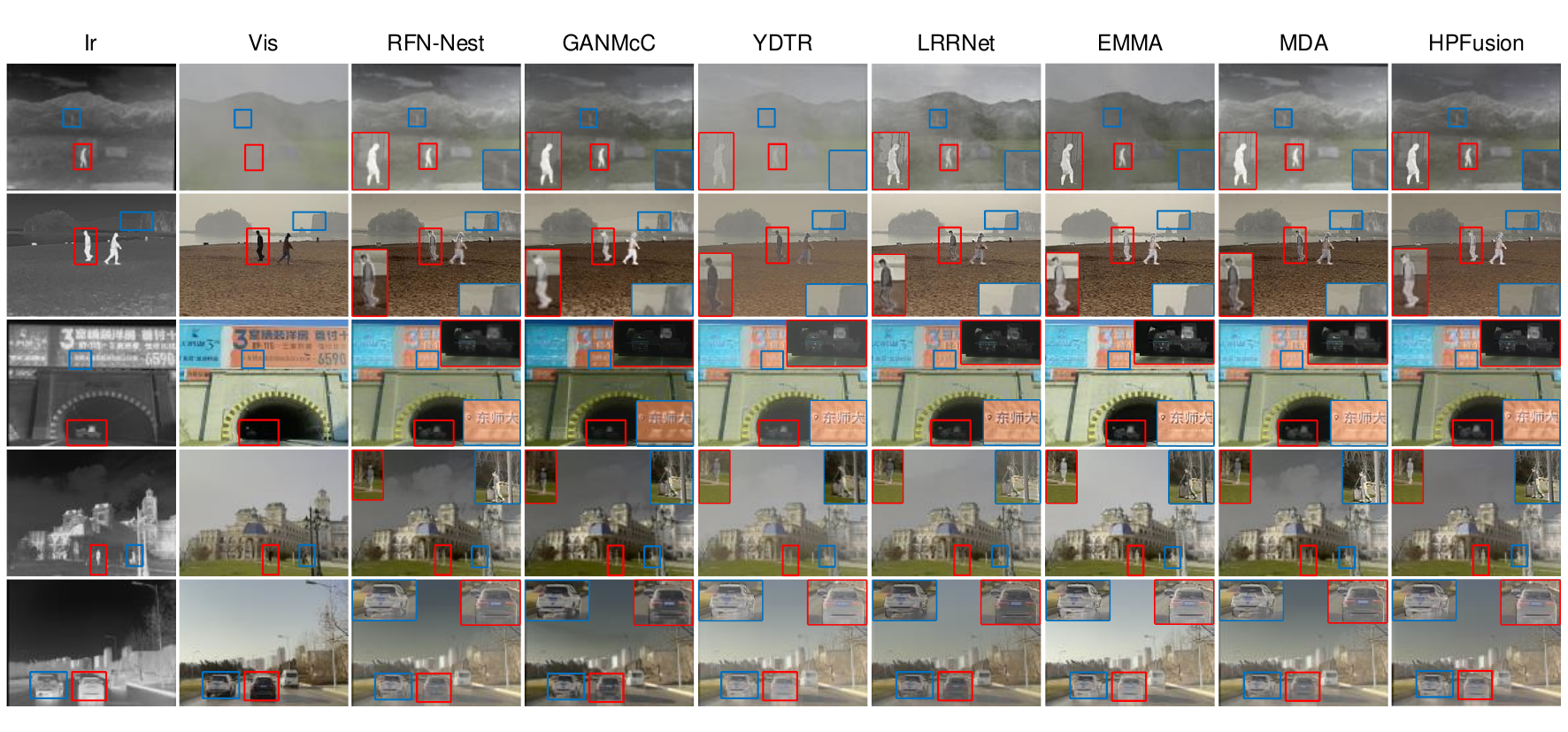}}
\caption{Qualitative comparison of our method with 6 state-of-the-art models on five infrared and visible image pairs of the $M^3FD$ dataset. The first and second columns are infrared and visible images, respectively. From the third to ninth columns are images fused by comparsion methods.}
\label{fig:compar}
\end{figure*}

The total loss function is composed of pixel intensity loss, detail loss and hierarchical semantic loss, which is as follows:
\begin{equation}
\label{eq:total_loss}
\begin{split} 
  L_{total} =& L_{image} + \beta \times L_{hier} \\
            =& L_{int}+ \alpha \times L_{detail} + \beta \times L_{hier}
\end{split} 
\end{equation}

We set values of $\alpha$ and $\beta$ to 4 and 1, respectively.

\section{Experiments}
In this section, we demonstrate the effectiveness of our HPFusion on preserving thermal and detail information. We train and test our on the $M^3FD$ dataset\cite{liu2022target}, with 4200 image pairs in the training set and 300 image pairs in the test set. We train our network for 100 epochs with Adam optimizer and batch size is set to 8. All the images during the training phase are resized to $224 \times 224$. Two NVIDIA GeForce RTX 3090 GPUs are used in the training. One for the inference of LLaVA and the other for the training of the fusion network.

\subsection{Qualitative Experiments}
We choose six state-of-the-art image fusion methods to compare the performance of preserving salient targets and detail information, which are RFN-Nest\cite{li2021rfn}, LRRNet\cite{li2023lrrnet}, YDTR\cite{tang2022ydtr}, GANMcC\cite{ma2020ganmcc}, EMMA\cite{zhao2024equivariant} and our baseline MDA\cite{yang2024multi}.  Qualitative results are shown in Fig.~\ref{fig:compar}. The first two columns are source infrared and visible images, and the red and blue boxes in the fusion results from the third to ninth columns enlarge the details in the fused images. Some thermal targets exhibit low intensity in visible images, such as humans in the second rows and vehicles in the last rows. Compared with LRRNet and YDTR, our HPFusion can highlight the saliency of these targets, meanwhile keeping the contrast of visible images. Additionally, the detail information is effectively preserved in our results, like the sharp edges of targets and background, which is damaged in fused images generated by the GANMcC. 

\subsection{Quantitative Experiments}
Five quantitative metrics are employed for evaluation, which are mean squared error (MSE), structural similarity index metric (SSIM), correlation coefficient (CC), peak signal-to-noise ratio (PSNR) and edge retentiveness ($Q^{AB/F}$). These metrics provide a comprehensive assessment of the retention of pixel intensity and edge information transferred from the source images. Specifically, a lower MSE indicates a higher degree of preserving pixel intensity, whereas the other four metrics indicate the opposite trend. Comparison results are shown in upper part of Table~\ref{comparsion}, where the method proposed achieves the best performance in four metrics, indicating the effective preservation of thermal and detail information.

\subsection{Ablation Studies}
We investigate the effectiveness of HPM and hierarchical semantic loss in our ablation studies. Quantitative results are shown in the lower part of Table~\ref{comparsion}. Our HPFusion consisting of HPM and $L_{hier}$ achieves best performance across four metrics, which demonstrates the superiority of HPFusion in balancing the thermal saliency and texture details.

\section{Conclusion}
In this work, we propose an infrared and visible image fusion method incorporating hierarchical human perception by leveraging Large Vision-Language Model. We design four questions and use LLaVA to generate the texts of answers, which are subsequently encoded into textual embeddings by CLIP for guiding fusion and optimization. Experiments show that our fusion method can effectively preserve thermal and detail information, thereby enhancing human comprehension of the context of the image pair.
\begin{table}[htbp]
\caption{Quantitative comparison with state-of-the-art methods and ablation studies. The best performance are shown in \textbf{bold}, and the second and third best performance are shown in \textcolor[RGB]{255,0,0}{red} and \textcolor[RGB]{0,0,255}{blue}, respectively}
\begin{center}
\begin{tabular}{|c|c|c|c|c|c|c|}
\hline
\multicolumn{2}{|c|}{\textbf{Method}} & MSE$\downarrow$ & SSIM$\uparrow$ & PSNR$\uparrow$ & CC$\uparrow$ & $Q^{AB/F}\uparrow$\\
\hline
\multicolumn{2}{|c|}{RFN-Nest} & \textcolor[RGB]{0,0,255}{0.034} & 0.397 & \textcolor[RGB]{255,0,0}{63.373} & \textcolor[RGB]{0,0,255}{0.572} & 0.406 \\
\hline
\multicolumn{2}{|c|}{GANMcC} & 0.037 & 0.391 & \textcolor[RGB]{0,0,255}{62.956} & 0.571 & 0.268  \\
\hline
\multicolumn{2}{|c|}{YDTR} & 0.044 & \textcolor[RGB]{255,0,0}{0.471} & 62.728 & 0.554 & 0.478  \\
\hline
\multicolumn{2}{|c|}{LRRNet} & 0.039 & 0.388 & 62.952 & 0.541 & 0.498  \\
\hline
\multicolumn{2}{|c|}{EMMA} & 0.057 & \textcolor[RGB]{0,0,255}{0.451} & 61.686 & 0.502 & \textbf{0.592}  \\
\hline
\multicolumn{2}{|c|}{MDA} & \textcolor[RGB]{255,0,0}{0.033} & 0.438 & 62.517 & \textcolor[RGB]{255,0,0}{0.585} & \textcolor[RGB]{0,0,255}{0.487}  \\
\hline
\multicolumn{2}{|c|}{HPFusion} & \textbf{0.032} & \textbf{0.500} & \textbf{63.794} & \textbf{0.595} & \textcolor[RGB]{255,0,0}{0.505}  \\
\hline
\multicolumn{7}{|c|}{ \textbf{Ablation Studies}} \\
\hline
\textbf{HPM}& $\boldsymbol{L_{hier}}$ & MSE$\downarrow$ & SSIM$\uparrow$ & PSNR$\uparrow$ & CC$\uparrow$ & $Q^{AB/F}\uparrow$\\
\hline
$\times$ & $\surd$ & \textcolor[RGB]{255,0,0}{0.033} & \textcolor[RGB]{255,0,0}{0.489} & \textcolor[RGB]{255,0,0}{63.677} & \textcolor[RGB]{255,0,0}{0.574} & \textbf{0.523} \\
\hline
$\surd$ & $\times$ & \textcolor[RGB]{0,0,255}{0.058} & \textcolor[RGB]{0,0,255}{0.467} & \textcolor[RGB]{0,0,255}{61.507} & \textcolor[RGB]{0,0,255}{0.543} & \textcolor[RGB]{0,0,255}{0.502}  \\
\hline
$\surd$ & $\surd$ & \textbf{0.032} & \textbf{0.500} & \textbf{63.794} & \textbf{0.580} & \textcolor[RGB]{255,0,0}{0.505}  \\
\hline
\end{tabular}
\label{comparsion}
\end{center}
\end{table}

% \begin{table}[htbp]
% \caption{Quantitative comparison of ablation studies. The best performance are shown in \textbf{bold}, and the second and third best performance are shown in \textcolor[RGB]{255,0,0}{red} and \textcolor[RGB]{0,0,255}{blue}, respectively}
% \begin{center}
% \begin{tabular}{|c|c|c|c|c|c|c|}
% \hline
% \textbf{HPM}& $\boldsymbol{L_{hier}}$ & MSE$\downarrow$ & SSIM$\uparrow$ & PSNR$\uparrow$ & CC$\uparrow$ & $Q^{AB/F}\uparrow$\\
% \hline
% $\times$ & $\surd$ & 111 & 111 & 111 & 111 & 111 \\
% \hline
% $\surd$ & $\times$ & 111 & 111 & 111 & 111 & 111  \\
% \hline
% $\surd$ & $\surd$ & 0.032 & 0.500 & 63.794 & 0.580 & 0.505  \\
% \hline
% \end{tabular}
% \label{ablation}
% \end{center}
% \end{table}

\section*{Acknowledgment}

This work is supported by the National Natural Science Foundation of China under Grants U21A20514, 62176195, 62441601 and 62036007.

\bibliographystyle{IEEEtran}
\bibliography{refs}

\begin{thebibliography}{10}
\providecommand{\url}[1]{#1}
\csname url@samestyle\endcsname
\providecommand{\newblock}{\relax}
\providecommand{\bibinfo}[2]{#2}
\providecommand{\BIBentrySTDinterwordspacing}{\spaceskip=0pt\relax}
\providecommand{\BIBentryALTinterwordstretchfactor}{4}
\providecommand{\BIBentryALTinterwordspacing}{\spaceskip=\fontdimen2\font plus
\BIBentryALTinterwordstretchfactor\fontdimen3\font minus \fontdimen4\font\relax}
\providecommand{\BIBforeignlanguage}[2]{{%
\expandafter\ifx\csname l@#1\endcsname\relax
\typeout{** WARNING: IEEEtran.bst: No hyphenation pattern has been}%
\typeout{** loaded for the language `#1'. Using the pattern for}%
\typeout{** the default language instead.}%
\else
\language=\csname l@#1\endcsname
\fi
#2}}
\providecommand{\BIBdecl}{\relax}
\BIBdecl

\bibitem{li2017pixel}
S.~Li, X.~Kang, L.~Fang, J.~Hu, and H.~Yin, ``Pixel-level image fusion: A survey of the state of the art,'' \emph{information Fusion}, vol.~33, pp. 100--112, 2017.

\bibitem{ma2019infrared}
J.~Ma, Y.~Ma, and C.~Li, ``Infrared and visible image fusion methods and applications: A survey,'' \emph{Information fusion}, vol.~45, pp. 153--178, 2019.

\bibitem{li2018densefuse}
H.~Li and X.-J. Wu, ``Densefuse: A fusion approach to infrared and visible images,'' \emph{IEEE Transactions on Image Processing}, vol.~28, no.~5, pp. 2614--2623, 2018.

\bibitem{tang2022image}
L.~Tang, J.~Yuan, and J.~Ma, ``Image fusion in the loop of high-level vision tasks: A semantic-aware real-time infrared and visible image fusion network,'' \emph{Information Fusion}, vol.~82, pp. 28--42, 2022.

\bibitem{liu2023multi}
J.~Liu, Z.~Liu, G.~Wu, L.~Ma, R.~Liu, W.~Zhong, Z.~Luo, and X.~Fan, ``Multi-interactive feature learning and a full-time multi-modality benchmark for image fusion and segmentation,'' in \emph{Proceedings of the IEEE/CVF international conference on computer vision}, 2023, pp. 8115--8124.

\bibitem{liang2023iterative}
Z.~Liang, C.~Li, S.~Zhou, R.~Feng, and C.~C. Loy, ``Iterative prompt learning for unsupervised backlit image enhancement,'' in \emph{Proceedings of the IEEE/CVF International Conference on Computer Vision}, 2023, pp. 8094--8103.

\bibitem{yang2023implicit}
S.~Yang, M.~Ding, Y.~Wu, Z.~Li, and J.~Zhang, ``Implicit neural representation for cooperative low-light image enhancement,'' in \emph{Proceedings of the IEEE/CVF international conference on computer vision}, 2023, pp. 12\,918--12\,927.

\bibitem{sun2024coser}
H.~Sun, W.~Li, J.~Liu, H.~Chen, R.~Pei, X.~Zou, Y.~Yan, and Y.~Yang, ``Coser: Bridging image and language for cognitive super-resolution,'' in \emph{Proceedings of the IEEE/CVF Conference on Computer Vision and Pattern Recognition}, 2024, pp. 25\,868--25\,878.

\bibitem{yu2024scaling}
F.~Yu, J.~Gu, Z.~Li, J.~Hu, X.~Kong, X.~Wang, J.~He, Y.~Qiao, and C.~Dong, ``Scaling up to excellence: Practicing model scaling for photo-realistic image restoration in the wild,'' in \emph{Proceedings of the IEEE/CVF Conference on Computer Vision and Pattern Recognition}, 2024, pp. 25\,669--25\,680.

\bibitem{liu2024visual}
H.~Liu, C.~Li, Q.~Wu, and Y.~J. Lee, ``Visual instruction tuning,'' \emph{Advances in neural information processing systems}, vol.~36, 2024.

\bibitem{radford2021learning}
A.~Radford, J.~W. Kim, C.~Hallacy, A.~Ramesh, G.~Goh, S.~Agarwal, G.~Sastry, A.~Askell, P.~Mishkin, J.~Clark \emph{et~al.}, ``Learning transferable visual models from natural language supervision,'' in \emph{International conference on machine learning}.\hskip 1em plus 0.5em minus 0.4em\relax PMLR, 2021, pp. 8748--8763.

\bibitem{veshki2022coupled}
F.~G. Veshki and S.~A. Vorobyov, ``Coupled feature learning via structured convolutional sparse coding for multimodal image fusion,'' in \emph{ICASSP 2022-2022 IEEE International Conference on Acoustics, Speech and Signal Processing (ICASSP)}.\hskip 1em plus 0.5em minus 0.4em\relax IEEE, 2022, pp. 2500--2504.

\bibitem{zhou2016perceptual}
Z.~Zhou, B.~Wang, S.~Li, and M.~Dong, ``Perceptual fusion of infrared and visible images through a hybrid multi-scale decomposition with gaussian and bilateral filters,'' \emph{Information fusion}, vol.~30, pp. 15--26, 2016.

\bibitem{ma2019fusiongan}
J.~Ma, W.~Yu, P.~Liang, C.~Li, and J.~Jiang, ``Fusiongan: A generative adversarial network for infrared and visible image fusion,'' \emph{Information fusion}, vol.~48, pp. 11--26, 2019.

\bibitem{liu2022target}
J.~Liu, X.~Fan, Z.~Huang, G.~Wu, R.~Liu, W.~Zhong, and Z.~Luo, ``Target-aware dual adversarial learning and a multi-scenario multi-modality benchmark to fuse infrared and visible for object detection,'' in \emph{Proceedings of the IEEE/CVF conference on computer vision and pattern recognition}, 2022, pp. 5802--5811.

\bibitem{zhao2023ddfm}
Z.~Zhao, H.~Bai, Y.~Zhu, J.~Zhang, S.~Xu, Y.~Zhang, K.~Zhang, D.~Meng, R.~Timofte, and L.~Van~Gool, ``Ddfm: denoising diffusion model for multi-modality image fusion,'' in \emph{Proceedings of the IEEE/CVF International Conference on Computer Vision}, 2023, pp. 8082--8093.

\bibitem{zhao2024image}
\BIBentryALTinterwordspacing
Z.~Zhao, L.~Deng, H.~Bai, Y.~Cui, Z.~Zhang, Y.~Zhang, H.~Qin, D.~Chen, J.~Zhang, P.~WANG, and L.~V. Gool, ``Image fusion via vision-language model,'' in \emph{Forty-first International Conference on Machine Learning}, 2024. [Online]. Available: \url{https://openreview.net/forum?id=eqY64Z1rsT}
\BIBentrySTDinterwordspacing

\bibitem{li2022blip}
J.~Li, D.~Li, C.~Xiong, and S.~Hoi, ``Blip: Bootstrapping language-image pre-training for unified vision-language understanding and generation,'' in \emph{International conference on machine learning}.\hskip 1em plus 0.5em minus 0.4em\relax PMLR, 2022, pp. 12\,888--12\,900.

\bibitem{yang2024multi}
G.~Yang, J.~Li, H.~Lei, and X.~Gao, ``A multi-scale information integration framework for infrared and visible image fusion,'' \emph{Neurocomputing}, vol. 600, p. 128116, 2024.

\bibitem{li2021rfn}
H.~Li, X.-J. Wu, and J.~Kittler, ``Rfn-nest: An end-to-end residual fusion network for infrared and visible images,'' \emph{Information Fusion}, vol.~73, pp. 72--86, 2021.

\bibitem{li2023lrrnet}
H.~Li, T.~Xu, X.-J. Wu, J.~Lu, and J.~Kittler, ``Lrrnet: A novel representation learning guided fusion network for infrared and visible images,'' \emph{IEEE transactions on pattern analysis and machine intelligence}, vol.~45, no.~9, pp. 11\,040--11\,052, 2023.

\bibitem{tang2022ydtr}
W.~Tang, F.~He, and Y.~Liu, ``Ydtr: Infrared and visible image fusion via y-shape dynamic transformer,'' \emph{IEEE Transactions on Multimedia}, vol.~25, pp. 5413--5428, 2022.

\bibitem{ma2020ganmcc}
J.~Ma, H.~Zhang, Z.~Shao, P.~Liang, and H.~Xu, ``Ganmcc: A generative adversarial network with multiclassification constraints for infrared and visible image fusion,'' \emph{IEEE Transactions on Instrumentation and Measurement}, vol.~70, pp. 1--14, 2020.

\bibitem{zhao2024equivariant}
Z.~Zhao, H.~Bai, J.~Zhang, Y.~Zhang, K.~Zhang, S.~Xu, D.~Chen, R.~Timofte, and L.~Van~Gool, ``Equivariant multi-modality image fusion,'' in \emph{Proceedings of the IEEE/CVF Conference on Computer Vision and Pattern Recognition}, 2024, pp. 25\,912--25\,921.

\end{thebibliography}

\end{document}